\pdfoutput=1
\documentclass[11pt]{article}

\usepackage{main}
\usepackage{multicol}
\usepackage{multirow}
\usepackage{graphicx}
\usepackage{times}
\usepackage{latexsym}
\usepackage[utf8]{inputenc}
\usepackage{microtype}
\usepackage{inconsolata}
\usepackage{amsmath}
\usepackage{amssymb}
\usepackage{makecell}
\usepackage{tikz}
\usepackage[font=small,labelfont=bf]{caption}

\newcommand\encircle[2][]{\tikz[overlay]\node[fill=blue!20,inner sep=2pt, anchor=text, rectangle, rounded corners=1.5mm,#1] {#2};\phantom{#2}}
\definecolor{myGreen}{RGB}{127,210,85}
\definecolor{myOrange}{RGB}{242,154,66}
\definecolor{myYellow}{RGB}{247,223,65}
\definecolor{myRed}{RGB}{232,80,43}
\definecolor{myViolet}{RGB}{162,57,102}
\definecolor{myBlue}{HTML}{4686f3}
\definecolor{myYellowv2}{HTML}{E6C802}
\definecolor{myOrangev2}{HTML}{ED8E55}
\definecolor{MyGreenv2}{HTML}{009B55}
\definecolor{MyRedv2}{HTML}{c22f2f}
\newcommand{\mwp}{\textsc{MWP}}  

\title{Large Language Models for Mathematical Reasoning: \\Progresses and Challenges}

\author{
  Janice Ahn\textsuperscript{\rm $\spadesuit$} \quad
  Rishu Verma\textsuperscript{\rm $\spadesuit$} \quad
  Renze Lou\textsuperscript{\rm $\spadesuit$} \quad
  Di Liu\textsuperscript{\rm $\diamondsuit$} \quad
  \\
  \textbf{Rui Zhang}\textsuperscript{\rm $\spadesuit$} \quad
  \and
  \textbf{Wenpeng Yin}\textsuperscript{\rm $\spadesuit$}
  \\
  \textsuperscript{\rm $\spadesuit$}The Pennsylvania State University
  \ 
  \textsuperscript{\rm $\diamondsuit$} Temple University
  \\
  {\texttt{\{jfa5672, wenpeng\}@psu.edu};}
  \ 
  { \texttt{diliu@temple.edu}}
}

\date{}

\begin{document}
\maketitle
\begin{abstract}
Mathematical reasoning serves as a cornerstone for assessing the fundamental cognitive capabilities of human intelligence. In recent times, there has been a notable surge in the development of Large Language Models (LLMs) geared towards the automated resolution of mathematical problems. However, the landscape of mathematical problem types is vast and varied, with LLM-oriented techniques undergoing evaluation across diverse datasets and settings. This diversity makes it challenging to discern the true advancements and obstacles within this burgeoning field. This survey endeavors to address four pivotal dimensions: i) a comprehensive exploration of the various mathematical problems and their corresponding datasets that have been investigated; ii) an examination of the spectrum of LLM-oriented techniques that have been proposed for mathematical problem-solving; iii) an overview of factors and concerns affecting LLMs in solving math; and iv) an elucidation of the persisting challenges within this domain. To the best of our knowledge, this survey stands as one of the first extensive examinations of the landscape of LLMs in the realm of mathematics, providing a holistic perspective on the current state, accomplishments, and future challenges in this rapidly evolving field.
\end{abstract}                     
\section{Introduction}
Mathematical reasoning is crucial to human intelligence, driving ongoing efforts in the AI community to autonomously tackle math challenges. This pursuit inherently calls for an augmentation of AI capabilities, delving into the intricate realms of textual comprehension, image interpretation, tabular analysis, symbolic manipulation, operational logic, and a nuanced grasp of world knowledge. As the AI landscape evolves, the endeavor to empower machines with a comprehensive understanding of diverse mathematical facets becomes not only a testament to technological prowess but also a pivotal stride towards achieving a more generalized and adept AI.

In recent times, the landscape of AI has been reshaped by the ascendancy of Large Language Models (LLMs) as formidable tools for automating intricate tasks. Notably, LLMs have proven to be potent assets in unraveling the nuances of mathematical problem-solving \cite{romera2023mathematical,DBLPImaniD023}. Their language capabilities fuel focused exploration in utilizing them for mathematical reasoning, uncovering fresh insights into the synergy between language and logic.

However, amid this progress, the current state of LLM-oriented research in mathematics presents a complex panorama. Diverse mathematical problem types pose a formidable challenge, exacerbated by the varied evaluation metrics, datasets, and settings employed in the assessment of LLM-oriented techniques \cite{DBLP07735,DBLPLu00WC23}. The lack of a unified framework hampers our ability to gauge the true extent of progress achieved and impedes a coherent understanding of the challenges that persist in this evolving field.

This survey endeavors to cast a spotlight on the multifaceted landscape of LLMs in the realm of mathematics. We plan to traverse four crucial dimensions: a meticulous exploration of math problem types and the datasets associated with them; an in-depth analysis of the evolving techniques employed by LLMs in mathematical problem-solving; an examination of factors that affect the LLMs solving math problems; and a critical discussion on the persisting challenges that loom over this burgeoning field. 

To our knowledge, this survey marks one of the first comprehensive examinations of LLMs specifically tailored for mathematics. By weaving together insights from various dimensions, we aim to provide a holistic understanding of the current state of affairs in LLM-driven mathematical reasoning, shedding light on achievements, challenges, and the uncharted territories that await exploration in this captivating intersection of language and logic.

\section{Related Work} 

To the best of our knowledge, the existing literature on summarizing mathematical research, particularly within the context of LLMs, remains limited. Notably, \newcite{DBLP04556} compared two ChatGPT versions (9-January-2023 and 30-January-2023) and GPT-4 for four math-related problems: producing proofs, filling holes in proofs, acting as a mathematical search engine and computation. More importantly, they summarized some insightful strategies regarding how LLMs can help mathematicians and advocated a more collaborative
approach, incorporating human expertise and LLM automation, for theorem proving. \citet{DBLP03109} conducted a comprehensive evaluation of LLMs, incorporating an examination of their performance in mathematical problem-solving, albeit with a relatively brief exploration of the mathematical field. Conversely, both \cite{DBLP07735} and \cite{DBLPLu00WC23} delved into the application of Deep Learning in the domain of mathematical reasoning.  Our work distinguishes itself on three fronts: firstly, we concentrate on LLMs, providing a more in-depth analysis of their various advancements; secondly, beyond merely reporting progress, we engage in a thorough discussion of the challenges inherent in this trajectory; and thirdly, we extend our scrutiny to encompass the perspective of mathematics pedagogy. In doing so, we contribute a nuanced perspective that seeks to broaden the understanding of LLMs in the context of mathematical research. 

The only work contemporaneous with ours is \cite{DBLP07622}. In comparison, our contribution lies in: i) not only introducing various methods but also paying more attention to various factors affecting model performance; ii) taking a broader perspective on the progress of LLM in the field of mathematics, elucidating not only from the AI perspective but also from the perspective of education. It emphasizes that the pursuit of model performance alone, while neglecting human factors, is something that needs attention.

\section{Math Problems \& Datasets}
This section concisely overviews prominent mathematical problem types and associated datasets, spanning \textsc{Arithmetic}, \textsc{Math Word Problems}, \textsc{Geometry},  \textsc{Automated Theorem Proving}, and \textsc{Math in vision context}.

\subsection{Arithmetic}
This category of problems entails pure mathematical operations and numerical manipulation, devoid of the need for the model to interpret text, images, or other contextual elements. An illustrative example is presented below, where ``$\mathcal{Q}$'' denotes questions and ``$\mathcal{A}$'' for answers.

\vspace{2mm}
\noindent
\fbox{%
 \begin{minipage}{0.95\linewidth}
   $\mathcal{Q}$: \texttt{21 + 97 } \\
   $\mathcal{A}$: \texttt{ 118} 
 \end{minipage}
}
\vspace{1mm}

The dataset \textsc{Math-140} \cite{DBLP02015} contains 401 arithmetic expressions for 17 groups.

\bgroup
\begin{table*}[!t]
    \centering
    \small
    \def\arraystretch{1.2}%
    \begin{tabular}{|c|p{5.5cm}|r|c|l|}
    \hline
       & \thead{\textbf{\textsc{Name}}} & \thead{\textbf{\textsc{Size}}} & 
       \thead{\textbf{\textsc{Level}}} &
       \thead{\textbf{\textsc{Note}}} \\ \hline
      \multirow{2}{*}{\rotatebox{90}{
      \begin{tabular}{c} Q-A \end{tabular}}} &
      \textsc{CMath} \cite{DBLP16636} & 1.7K & \encircle[fill=MyGreenv2, text=white]{E} & Chinese; grade 1-6 \\
      & \textsc{Sat-Math} \cite{DBLP06364} & 220 & \encircle[fill=myYellowv2, text=white]{H}& Multi-choice \\
    \hline
             \multirow{12}{*}{\rotatebox{90}{\begin{tabular}{c} Question-Equation-Answer\end{tabular}}} & \textsc{SVAMP} \cite{DBLPPatelBG21} & 1K& \encircle[fill=MyGreenv2, text=white]{E} & Three types of variations\\ 
             & \textsc{ASDiv} \cite{DBLPMiaoLS20} & 2.3K&  \encircle[fill=MyGreenv2, text=white]{E} & Problem type and grade level annotated\\
             & \textsc{MAWPS} \cite{DBLPKedziorski16} & 3.3K & \encircle[fill=MyGreenv2, text=white]{E} & Extension of \textsc{AddSub}, \textsc{MultiArith}, etc. \\
             & \textsc{ParaMAWPS} \cite{DBLP13899}& 16K & \encircle[fill=MyGreenv2, text=white]{E}  &  Paraphrased, adversarial \textsc{MAWPS}\\
             & \textsc{SingleEq} \cite{DBLPKoncel-Kedziorski15} & 508 & \encircle[fill=MyGreenv2, text=white]{E}& \\
             & \textsc{AddSub} \cite{DBLPHosseiniHEK14} & 395 & \encircle[fill=MyGreenv2, text=white]{E} & Only addition and subtraction \\
             & \textsc{MultiArith} \cite{DBLPRoyR15} & 600 & \encircle[fill=MyGreenv2, text=white]{E} & Multi-step reasoning\\
             & \textsc{Draw-1K} \cite{DBLPUpadhyayC17} & 1K & \encircle[fill=MyGreenv2, text=white]{E} & \\
             & \textsc{Math23K} \cite{DBLPWangLS17} & 23K & \encircle[fill=MyGreenv2, text=white]{E} & Chinese \\

             & \textsc{Ape210K} \cite{DBLP11506} & 210K & \encircle[fill=MyGreenv2, text=white]{E}& Chinese\\
             & \textsc{K6} \cite{DBLP03241} & 600 & \encircle[fill=MyGreenv2, text=white]{E} & Chinese;  grade 1-6\\
                          & \textsc{CM17K} \cite{DBLPQinLHTL20} & 17K & \encircle[fill=myBlue, text=white]{M}  
 \hspace{3pt}\encircle[fill=myYellowv2, text=white]{H} & Chinese; grade 6-12 \\

    \hline
             \multirow{12}{*}{\rotatebox{90}{\begin{tabular}{c} Question-Rationale-Answer\end{tabular}}} &               \textsc{Carp} \cite{zhang2023evaluating} &4.9K & \encircle[fill=myBlue, text=white]{M} & Chinese \\
             & \textsc{GSM8K} \cite{DBLP14168} & 8.5K & \encircle[fill=myBlue, text=white]{M} & Linguistically diverse \\
             &\textsc{MATH} \cite{DBLPHendrycksBKABTS21} &12.5K & \encircle[fill=myYellowv2, text=white]{H} & Problems are put into difficulty levels 1-5\\
             & \textsc{PRM800K} \cite{DBLP20050} & 12K & \encircle[fill=myYellowv2, text=white]{H} & \textsc{MATH} w/ step-wise labels\\
             & \textsc{MathQA} \cite{DBLPAminiGLKCH19} &37K& \encircle[fill=myOrangev2, text=white]{C} & GRE examinations; have quality concern\\
             & \textsc{AQuA} \cite{DBLPLingYDB17} & 100K & \encircle[fill=myOrangev2, text=white]{C} & GRE\&GMAT questions\\
             & \textsc{Arb} \cite{DBLP13692} & 105 & \encircle[fill=myOrangev2, text=white]{C} &  Contest problems and university math proof\\
             & \textsc{Ghosts} \cite{DBLP13867} & 709 & \encircle[fill=myOrangev2, text=white]{C} & \\
             & \textsc{TheoremQA-Math} \cite{DBLPChenYKLWMXWX23} & 442 & \encircle[fill=myOrangev2, text=white]{C} & Theorem as rationale\\
             & \textsc{LILA} \cite{DBLPMishraFLTWBRTSC22} & 132K & \encircle[fill=MyRedv2, text=white]{H}& Incorporates 20 existing datasets\\
             & \textsc{Math-Instruct} \cite{DBLP05653} & 260K & \encircle[fill=MyRedv2, text=white]{H}& Instruction-following style\\
             & \textsc{TabMWP} \cite{DBLPLu0CWZRCK23} & 38K & \encircle[fill=MyRedv2, text=white]{H}& Tabular MWP; below the College level\\
             \hline
    \end{tabular}
    \centering
    \caption{Datasets for Math Word Problems. \\ \encircle[fill=MyGreenv2, text=white]{E} = Elementary, \encircle[fill=myBlue, text=white]{M} = Middle School, \encircle[fill=myYellowv2, text=white]{H} = High School, \encircle[fill=myOrangev2, text=white]{C} = College, \encircle[fill=MyRedv2, text=white]{H} = Hybrid  }
    \label{tab:MWP}
\end{table*}
\egroup
\subsection{Math Word Problems}
\textsc{Math word problems} (\mwp) are mathematical exercises or scenarios presented in the form of written or verbal descriptions rather than straightforward equations in \textsc{Arithmetic}. These problems require individuals to decipher the information provided, identify relevant mathematical concepts, and formulate equations or expressions to solve the given problem. \mwp~often reflect real-world situations, allowing individuals to apply mathematical principles to practical contexts. Solving these problems typically involves critical thinking, problem-solving skills, and the application of mathematical operations to find a solution.

\mwp~invariably comprise a question ($\mathcal{Q}$) and its corresponding final answer ($\mathcal{A}$) (referred to as \emph{Question-Answer}). However, the presence or absence of additional clues can give rise to various versions of these problems. Variations may emerge based on factors such as the availability of an equation ($\mathcal{E}$; referred to as \emph{Question-Equation-Answer}) or the provision of a step-by-step rationale ($\mathcal{R}$; \emph{Question-Rationale-Answer}) to guide the problem-solving process. 

\paragraph{Question-Answer.} The instance of this type of \mwp~consists of a question ($\mathcal{Q}$) and the final answer ($\mathcal{A}$), such as:

\noindent
\fbox{%
 \begin{minipage}{0.95\linewidth}
   $\mathcal{Q}$: \texttt{Lily received \$20 from her mum. After spending \$10 on a storybook and \$2.5 on a lollipop, how much money does she have left?} \\
   $\mathcal{A}$: \texttt{ \$7.5} 
 \end{minipage}
}

\pagebreak
\paragraph{Question-Equation-Answer.} Compared with \emph{Question-Answer}, this \mwp~type provides the equation solution, such as

\noindent
\fbox{%
 \begin{minipage}{0.95\linewidth}
   $\mathcal{Q}$: \texttt{Jack had 8 pens and Mary had 5 pens. Jack gave 3 pens to Mary. How many pens does Jack have now?} \\
   $\mathcal{E}$: $  8 - 3 $ \\
   $\mathcal{A}$: \texttt{ 5}  (optional) 
 \end{minipage}
}



\paragraph{Question-Rationale-Answer.} This type of \mwp~includes answers and reasoning paths, akin to the Chain-of-Thought method, which explicates reasoning steps rather than defining problem types \cite{DBLPWei0SBIXCLZ22}. The rationale guides correct problem-solving and serves as a valuable reference for model training, including fine-tuning and few-shot learning.

\noindent
\fbox{%
 \begin{minipage}{0.95\linewidth}
   $\mathcal{Q}$: \texttt{Beth bakes 4, or 2 dozen batches of cookies in a week. If these cookies are shared amongst 16 people equally, how many cookies does each person consume?} \\
   $\mathcal{R}$: \texttt{Beth bakes 4 2 dozen batches of cookies for a total of $4*2 = <<4*2=8>>$ 8 dozen cookies. There are 12 cookies in a dozen and she makes 8 dozen cookies for a total of $12*8 = <<12*8=96>>$ 96 cookies. She splits the 96 cookies equally amongst 16 people so they each eat $96/16 = <<96/16=6>>$ 6 cookies.}\\
   $\mathcal{A}$: \texttt{ 6} 
 \end{minipage}
}

Table \ref{tab:MWP} lists most datasets that are summarized in three categories: \emph{Question-Answer}, \emph{Question-Equation-Answer}, and \emph{Question-Rationale-Answer}. In addition to the above three \mwp~types of conventional styles, recent work studied \mwp~in given tables and even \mwp~generation.

\pagebreak
\paragraph{Tabular \mwp.} \textsc{TabMWP} \cite{DBLPLu0CWZRCK23}  is the first dataset to study \mwp~over tabular context on open domains and is the largest in terms of data size. Each problem in \textsc{TabMWP} is accompanied by a tabular context, which is represented in three formats: an image, a semi-structured text, and a structured table.

 \begin{table}[h]
     \centering
     \begin{tabular}{|l|c|}
     \hline
       \thead{\textbf{\textsc{beads}}} & \thead{\textbf{\textsc{\$/kilogram}}} \\ \hline
          heart-shaped  & 3  \\
          rectangular  & 2\\
          spherical  & 2\\
          oval  & 2\\\hline
     \end{tabular}
     \caption{Table for the tabular \mwp~example.}
     \label{tab:tabmwp}
 \end{table}
\vspace{2mm}
\noindent
\fbox{%
 \begin{minipage}{0.95\linewidth}
 $\mathcal{T}$: Table \ref{tab:tabmwp}\\
   $\mathcal{Q}$: \texttt{Henrik bought 2.5 kilograms of oval beads. How much did he spend? (Unit: \$)} \\
   $\mathcal{A}$: \texttt{ 5} 
 \end{minipage}
}

\paragraph{\mwp~Generation.} Instead of deriving the answer for a given math question, this type of mathematical reasoning tries to generate \mwp~questions. For example, \citet{DBLP0001LB21} fine-tuned GPT-2 \cite{radford2019language} on equation-to-\mwp~instances for \mwp~generation. The effectiveness of GPT-3's question-generation capabilities was assessed by \citet{DBLPZongK23}, who instructed the model to generate a question similar to a provided \mwp~question. \citet{DBLP01991} analyzed a group of LLMs (GPT-4, GPT-3.5, PaLM-2 \cite{DBLP10403}, and LLaMa \cite{DBLP13971}), and found a significant drop in accuracy for backward reasoning compared to forward reasoning. \citet{DBLPNorbergAFLWM023} used GPT-4 to rewrite human-written \mwp, reporting optimal readability, lexical diversity, and cohesion scores, although GPT-4 rewrites incorporated more low-frequency words.


\subsection{Geometry}
Compared with \mwp, \textsc{geometry} problems involve a distinct set of challenges. While \mwp~often requires logical reasoning and arithmetic operations, geometry problems demand a spatial understanding of shapes, sizes, and their interrelationships. Solving geometry problems typically entails applying geometric principles, theorems, and formulas to analyze and deduce properties of geometric figures. Furthermore, current geometry approaches mainly rely on symbolic methods and predefined search heuristics, highlighting the specialized strategies required in this domain \cite{AlphaGeometryTrinh2024}. This contrast in problem-solving approaches highlights the multifaceted nature of mathematical challenges and the varied skill sets required in different mathematical domains. An example can be seen as follows and Table \ref{tab:geometry} lists mainstream datasets. 

\begin{figure}[!h]
\centering
\includegraphics[width=0.2\textwidth]{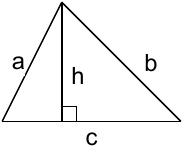}
\vspace{-5mm} 
\end{figure}
\noindent
\fbox{%
 \begin{minipage}{0.95\linewidth}
   $\mathcal{Q}$: \texttt{$a$=7 inches; $b$=24 inches; $c$=25 inches; $h$=6.72 inches; What is its area? (Unit: square inches)} \\
   $\mathcal{A}$: \texttt{84} 
 \end{minipage}
}

\begin{table}[t]
    \centering
    \begin{tabular}{|l|r|}
    \hline
       \thead{\textbf{\textsc{Name}}} & \thead{\textbf{\textsc{Size}}} \\ \hline
         \textsc{GeoShader} \cite{DBLPAlvinGMM17} & 102  \\
         \textsc{GeoS} \cite{DBLPSeoHFEM15} & 186  \\
         \textsc{GeoS++} \cite{DBLPSachanDX17} & 1.4K  \\
         \textsc{GeoS-OS} \cite{DBLPSachanX17} & 2.2K  \\
         \textsc{Geometry3K} \cite{DBLPLuGJQHLZ20} & 3K  \\
          \textsc{GeoQA} \cite{DBLPChenTQLLXL21} & 5K  \\
          \textsc{UniGeo} \cite{DBLPChenLQLLCL22} & 14.5K  \\ \hline
    \end{tabular}
    \caption{Geometry datasets}
    \label{tab:geometry}
\end{table}

\subsection{Automated theorem proving}
In the specialized area of Automated Theorem Proving (ATP), the inherent challenges are unique and encompass a wide spectrum, akin to those found in distinct mathematical fields. ATP's core focus is on autonomously constructing proofs for specified conjectures, requiring a blend of logical analysis and a profound grasp of formal languages, supported by an extensive knowledge base. Its application is crucial in areas like the validation and development of both software and hardware systems.

For example, the \textsc{MiniF2F} dataset \cite{zheng2022minif2f} stands out in ATP, featuring a series of complex Olympiad-level mathematical problems, designed to evaluate theorem-proving systems including Metamath \cite{DBLP12284}, Lean \cite{DBLPHanRWAP22}, and Isabelle \cite{wenzel2008isabelle}. In a similar vein, the HOList benchmark \cite{bansal2019holist}, with its comprehensive array of theorem statements from various corpora, sets a sequential proving challenge for ATP systems, where each theorem must be proved using only the lemmas preceding it. Additionally, the \textsc{CoqGym} dataset \cite{yang2019learning} provides a broad ATP environment, showcasing a rich collection of more than 71,000 proofs penned by humans, all within the framework of the Coq proof assistant. These datasets illustrate the diverse methodologies and skillsets necessary in ATP, reflecting the multifaceted nature of solving mathematical problems.

\subsection{Math in vision-language context}
\textsc{ChartQA} \cite{DBLPMasryLTJH22}, with 9.6K
human-written questions and 23.1K model-generated questions have explored a variety of complex reasoning questions that
involve several logical and arithmetic operations over charts. \textsc{MathVista} \cite{DBLP02255}: size: 6K; it features seven types of mathematical reasoning: algebraic reasoning, arithmetic reasoning, geometry reasoning, logical reasoning, numeric common sense, scientific reasoning, and statistical reasoning. In addition, fine-grained metadata are available, including question type, answer type, language, source, category, task, grade level, and visual context.

\section{Methodologies} 

We summarize these methods into three progressive levels: i) Prompting frozen LLMs, ii) Strategies enhancing frozen LLMs, and iii) Fine-tuning LLMs.

\subsection{Prompting frozen LLMs}

We organize prior work by typical LLMs.

\paragraph{GPT-3.} \citet{DBLPZongK23} evaluated the use of GPT-3, a 175B parameter transformer model for three related challenges pertaining to math word problems: i) classifying word problems, ii) extracting equations from word problems, and iii) generating word problems.

\paragraph{ChatGPT.} \citet{DBLPShakarianKNM23} reported the first independent evaluation of ChatGPT on \mwp, and found that ChatGPT’s performance changes dramatically based on the
requirement to show its work. \citet{cheng-zhang-2023-analyzing} assessed ChatGPT, OpenAI’s latest conversational chatbot and LLM, on its performance in elementary-grade arithmetic and logic problems, and found that ChatGPT performed better than previous models such as InstructGPT \cite{DBLPOuyang0JAWMZASR22} and Minerva \cite{lewkowycz2022solving}.

\paragraph{GPT-4.} \citet{DBLP01337}  adapted and evaluated several existing prompting methods to the usage of GPT-4, including a vanilla prompt, Program-of-Thoughts prompt \cite{DBLP12588}, and Program Synthesis prompt \cite{drori2022neural}. The study by \citet{DBLP10677} investigated the capability of GPT-4 to actively engage in math-oriented brainstorming sessions. This includes tasks like identifying new research problems, refining problem formulations, and suggesting potential methods or unconventional solutions, all achieved through iterative ideation with a human partner—a common practice in collaborative brainstorming with other professionals.

\paragraph{GPT4V \& Bard.} \citet{DBLP02255}  presented \textsc{MathVista}, a benchmark of evaluating mathematical reasoning in visual context, conducted a comprehensive, quantitative evaluation of three LLMs (i.e, ChatGPT, GPT-4, Claude-2 \cite{DBLP05862}), two proprietary large multimodal models (LMMs) (i.e., GPT4V, Bard), and seven open-source LMMs, with Chain-of-Thought   and Program-of-Thought.

\paragraph{Multiple.} \citet{DBLP16636} evaluated a variety of popular LLMs, including both commercial and open-source options, aiming to provide a benchmark tool for assessing the following question: to what grade level of Chinese elementary school math do the abilities of popular LLMs correspond?

\subsection{Strategies enhancing frozen LLMs}

\paragraph{Preprocessing the math question.} \citet{DBLPAnLG23} explored ChatGPT for the dataset \textsc{SVAMP}   and observed that substituting numerical expressions with English expressions can elevate the performance.

\paragraph{More advanced prompts.}  Chain-of-thought \cite{DBLPWei0SBIXCLZ22}, the first time to steer the LLMs to do \textbf{step-by-step math reasoning},  Self-Consistency \cite{DBLP0002WSLCNCZ23} tried multiple Chain-of-Thought reasoning paths and leverage the \textbf{consistency} mechanism to discover a more probable answer. \citet{DBLP07921} proposed a novel and effective prompting method, explicit code-based self-verification, to further boost the mathematical reasoning potential of GPT-4 Code Interpreter. This method employs a zero-shot prompt on GPT-4 Code Interpreter to encourage it to use code to \textbf{self-verify} its answers.

\paragraph{Using external tool.} \citet{DBLP13078} employed an external tool, specifically the Python REPL, to correct errors in Chain-of-Thought. Their demonstration highlighted that integrating Chain-of-Thought and Python REPL using a markup language improves the reasoning capabilities of ChatGPT. In a related context, \citet{DBLP09102} introduced an approach that merges an LLM, Codex \cite{DBLP03374}, capable of progressively formalizing word problems into variables and equations, with an external symbolic solver adept at solving the generated equations. Program-of-Thought \cite{DBLP12588} separates the computational aspect from the reasoning by utilizing a Language Model (primarily Codex) to articulate the reasoning procedure as a program. The actual computation is delegated to an external computer, responsible for executing the generated programs to arrive at the desired answer.

\paragraph{Improving the whole interaction.} \citet{DBLP01337} introduced MathChat, a conversational framework designed for chat-based LLMs. In this framework, math problems from the \textsc{MATH} dataset are resolved through a simulated conversation between the model and a user proxy agent.

\paragraph{Considering more comprehensive factors in evaluation.} While accuracy is crucial in evaluating LLMs for math problem-solving, it shouldn't be the sole metric. Other important dimensions include: i) \textbf{Confidence Provision:} \citet{DBLPImaniD023}'s "MathPromper" boosts LLM performance and confidence by generating algebraic expressions, providing diverse prompts, and evaluating consensus among multiple runs. ii) \textbf{Verifiable Explanations:} \citet{DBLPGaurS23} used concise, verifiable explanations to assess LLM reasoning, revealing their proficiency in zero-shot solving of symbolic \mwp and their ability to produce succinct explanations.

\subsection{Fine-tuning LLMs} 

\paragraph{Learning to select in-context examples.} As indicated by prior research, few-shot GPT-3's performance is susceptible to instability and may decline to near chance levels due to the reliance on in-context examples. This instability becomes more pronounced when dealing with intricate problems such as \textsc{TabMWP}. In addressing this issue, \citet{DBLPLu0CWZRCK23} introduced PROMPTPG, which can autonomously learn to select effective in-context examples through policy gradient interactions with the GPT-3 API, eliminating the need for manually designed heuristics.

\paragraph{Generating intermediate steps.}
\citet{DBLP00114} initiated the fine-tuning of decoder-only LLMs, ranging from 2M to 137B in size. Their approach involved training these models to solve integer addition and polynomial evaluation by generating intermediate computation steps into a designated ``scratchpad.'' In a related effort, \citet{DBLPZhang00FL23} introduced a fine-tuning strategy for GPT-2 or T5, enabling them to produce step-by-step solutions with a combination of textual and mathematical tokens leading to the final answer. Additionally, \citet{DBLP03241} applied a step-by-step strategy in fine-tuning a series of GLM models \cite{DBLPZengLDWL0YXZXTM23}, specifically tailored for solving distinct Chinese mathematical problems. Minerva, developed by \citet{lewkowycz2022solving}, enhances LLMs' ability to generate intermediate steps in complex math problems. Its fine-tuning of diverse datasets enables nuanced, step-by-step problem-solving, demonstrating advanced handling of intricate mathematical concepts.

\paragraph{Learning an answer verifier.} OpenAI researchers, per \citet{DBLP14168}, fine-tuned a GPT-3 model of 175B as a verifier, assigning probabilities to solution candidates. In exploring reexamination processes for \mwp~solving, \citet{DBLP09590} introduced Pseudo-Dual Learning, involving solving and reexamining modules. For \mwp~solution, \citet{DBLPZhuWZZ0GZY23} developed a cooperative reasoning-induced PLM, with GPT-J \cite{wang2021gpt} generating paths and DeBERTa-large \cite{DBLPHeLGC21} supervising evaluation. Google researchers, as per \citet{DBLP10047}, observed improved correctness in LLMs with multiple attempts, which hints that LLMs might generate correct solutions while struggling to differentiate between accurate and inaccurate ones. They sequentially fine-tuned their PaLM 2 model \cite{DBLP10403} as a solution generator, evaluator, and generator again.

\paragraph{Learning from enhanced dataset.} Emulating the error-driven learning process observed in human learning, \citet{DBLP20689} conducted fine-tuning on various open-source LLMs within the LLaMA \cite{DBLP13971}, LLaMA-2 \cite{DBLP09288}, CodeLLaMA \cite{DBLP12950}, WizardMath \cite{DBLP09583}, MetaMath \cite{DBLP12284}, and Llemma \cite{azerbayev2023llemma} families. This fine-tuning utilized mistake-correction data pairs generated by GPT-4.
To mitigate over-reliance on knowledge distillation from LLM teachers, \citet{DBLP07951} fine-tuned LLaMA-7B on existing mathematical problem datasets that exhibit diverse annotation styles. In a related approach, \citet{DBLP13899} demonstrated that training on linguistic variants of problem statements and implementing a voting mechanism for candidate predictions enhance the mathematical reasoning and overall robustness of the model.

\paragraph{Teacher-Student knowledge distillation.} \citet{DBLP14386} utilized GPT-3 to coach a more efficient MWP solver (RoBERTa-based encoder-decoder \cite{DBLP11692}). They shifted the focus from explaining existing exercises to identifying the student model's learning needs and generating new, tailored exercises. The resulting smaller LLM achieves competitive accuracy on the \textsc{SVAMP} dataset with significantly fewer parameters compared to state-of-the-art LLMs.

\paragraph{Finetuning on many datasets.} \citet{DBLPMishraFLTWBRTSC22} conducted fine-tuning on a series of GPT-Neo2.7B causal language models \cite{Black2021GPTNeoLS} using LILA, a composite of 20 existing math datasets. Similarly, \citet{DBLP05653} created ``MathInstruct'', a meticulously curated instruction tuning dataset. Comprising 13 math datasets with intermediate Chain-of-Thought and Program-of-Thought rationales, this dataset was used to fine-tune Llama \cite{DBLP13971, DBLP09288, DBLP12950} models across different scales. The resulting models demonstrate unprecedented potential in cross-dataset generalization.

\paragraph{Math solver ensemble.}
\citet{DBLPYaoZW23} incorporated a problem typing subtask that combines the strengths of the tree-based solver and the LLM solver (ChatGLM-6B \cite{DBLPZengLDWL0YXZXTM23}).


\section{Analysis}

\subsection{LLMs's robustness in math}

\citet{DBLPPatelBG21} provided strong evidence that the pre-LLM \mwp~solvers, mostly LSTM-equipped encoder-decoder models,  rely on shallow heuristics to achieve high performance on some simple benchmark datasets, then introduced a more challenging dataset, \textsc{SVAMP}, created by applying carefully chosen variations over examples sampled from preceding datasets. \citet{DBLPStolfoJSSS23} observed that, among non-instruction-tuned LLMs, the larger ones tend to be more sensitive to changes in the ground-truth result of a \mwp, but not necessarily more robust. However,  a different behavior exists in the instruction-tuned GPT-3 models, which show a remarkable improvement in both sensitivity and robustness, although the robustness reduces when problems get more complicated. \citet{DBLP16636}  assessed the robustness of several top-performing LLMs by augmenting the original problems in the curated \textsc{CMATH} dataset with distracting information. Their findings reveal that GPT-4 can maintain robustness while other models fail.

\citet{DBLP01686} proposed a new dataset \textsc{RobustMath} to evaluate the robustness of LLMs in math-solving ability. Extensive experiments show that (i)  Adversarial samples from higher-accuracy LLMs are also effective for attacking LLMs with lower accuracy; (ii) Complex MWPs (such as more solving steps, longer text, more numbers) are more vulnerable to attack; (iii) We can improve the robustness of LLMs by using adversarial samples in few-shot prompts.

\subsection{Factors in influencing LLMs in math}

The comprehensive evaluation conducted by \citet{DBLP02015} encompasses OpenAI's GPT series, including GPT-4, ChatGPT2, and GPT-3.5, along with various open-source LLMs. This analysis methodically examines the elements that impact the arithmetic skills of LLMs, covering aspects such as tokenization, pre-training, prompting techniques, interpolation and extrapolation, scaling laws, Chain of Thought (COT), and In-Context Learning (ICL).

\paragraph{Tokenization.} This research underscores tokenization's critical role in LLMs' arithmetic performance \cite{DBLP02015}. Models like T5, lacking specialized tokenization for arithmetic, are less effective than those with advanced methods, such as Galactica \cite{DBLP09085} and LLaMA, which show superior accuracy in arithmetic tasks. This indicates that token frequency in pre-training and the method of tokenization are key to arithmetic proficiency.

\paragraph{Pre-training Corpus.} Enhanced arithmetic skills in LLMs correlate with the inclusion of code and LATEX in pre-training data \cite{DBLP02015}. Galactica, heavily utilizing LATEX, excels in arithmetic tasks, while models like Code-DaVinci-002, better at reasoning, lags in arithmetic, highlighting a distinction between arithmetic and reasoning skills. 

\paragraph{Prompts.} The nature of input prompts greatly affects LLMs' arithmetic performance~\cite{liu2023pre,lou2023prompt}. Without prompts, performance drops \cite{DBLP02015}. Models like ChatGPT, which respond well to instructional system-level messages, demonstrate the importance of prompt type. Instruction tuning in pre-training also emerges as a significant factor \cite{DBLP05653}. 

\paragraph{Model Scale.} There's a noted correlation between parameter count and arithmetic capability in LLMs \cite{DBLP02015}. Larger models generally perform better, but a performance plateau is observed, as shown by Galactica's similar outcomes at 30B and 120B parameters. However, this doesn't always mean superior performance, with smaller models like ChatGPT occasionally outperforming larger ones.



\subsection{Perspectives of mathematics pedagogy}
While machine learning emphasizes LLMs' problem-solving abilities in mathematics, in practical education, their primary role is to aid learning. Thus, the focus shifts from mere mathematical performance to a crucial consideration of LLMs' understanding of students' needs, capabilities, and learning methods.

\paragraph{Advantages of deploying LLMs in math education.} Educators have observed the following benefits of leveraging LLMs for math education. (i) \emph{LLMs foster critical thinking and problem-solving skills}, as they provide comprehensive solutions and promote rigorous error analysis \cite{MatzDoukMoun2023ny}; (ii) \emph{Educators and students prefer LLM-generated hints} because of their detailed, sequential format and clear, coherent narratives \cite{Gattupalli2023Exploring}; (iii) \emph{LLMs introduce a conversational style in problem-solving}, an invaluable asset in math education \cite{Gattupalli2023Exploring}; (iv) The impact of LLMs extends \emph{beyond mere computational assistance}, offering deep insights and understanding \emph{spanning diverse disciplines} like Algebra, Calculus, and Statistics \cite{Rane2023Enhancing}.

\paragraph{Disadvantages of deploying LLMs in math education.} (i) \emph{Potential for misinterpretation.} Misinterpretation of students' queries or errors in providing explanations by LLMs could lead to confusion. Inaccurate responses might result in the reinforcement of misconceptions, impacting the quality of education \cite{DBLP13615}. (ii) \emph{Limited understanding of individual learning styles.} LLMs may struggle to cater to diverse learning styles, as they primarily rely on algorithms and might not fully grasp the unique needs of each student. Some learners may benefit more from hands-on activities or visual aids that LLMs may not adequately address.  \citet{Gattupalli2023Exploring} proposed that hints produced by GPT-4 could be excessively intricate for younger students who have shorter attention spans. (iii) \emph{Privacy and data security issues}. Deploying LLMs involves collecting and analyzing substantial amounts of student data. Privacy concerns may arise if proper measures are not in place to safeguard this data from unauthorized access or misuse.

\section{Challenges}

\paragraph{Data-driven \& limited generalization.} The prevailing trend in current research revolves around the curation of extensive datasets. Despite this emphasis, there is a noticeable lack of robust generalization across various datasets, grade levels, and types of math problems. Examining how humans acquire math-solving skills suggests that machines may need to embrace continual learning to enhance their capabilities.

\paragraph{LLMs' brittleness in math reasoning.} The fragility of LLMs in mathematical reasoning is evident across three dimensions. Firstly, when presented with questions expressed in varying textual forms (comprising words and numbers), LLMs exhibit inconsistent performance. Secondly, for identical questions, an LLM may yield different final answers through distinct reasoning paths during multiple trials. Lastly, pre-trained math-oriented LLMs are susceptible to attacks from adversarial inputs, highlighting their vulnerability in the face of manipulated data.

\paragraph{Human-oriented math interpretation.}
The current LLM-oriented math reasoning, such as chain-of-thoughts, does not take into account the needs and comprehension abilities of users, such as students. As an example, \citet{DBLP13615} discovered that GPT-3.5 had a tendency to misinterpret students' questions in the conversation, resulting in a failure to deliver adaptive feedback. Additionally, research conducted by \citet{Gattupalli2023Exploring} revealed that GPT-4 frequently overlooks the practical comprehension abilities of younger students. It tends to generate overly intricate hints that even confuse those students. Consequently, there is a pressing need for increased AI research that actively incorporates human factors into its design, ensuring future developments align more closely with the nuanced requirements of users.
\section{Conclusion}

This survey on LLMs for Mathematics delves into various aspects of LLMs in mathematical reasoning, including their capabilities and limitations. The paper discusses different types of math problems, datasets,  and the persisting challenges in the domain. It highlights the advancements in LLMs, their application in educational settings, and the need for a human-centric approach in math education. We hope this paper will guide and inspire future research in the LLM community, fostering further advancements and practical applications in diverse mathematical contexts.


\bibliography{anthology, custom}
\bibliographystyle{acl_natbib}



\end{document}